\pdfoutput=1

\documentclass[11pt]{article}

\usepackage[]{EMNLP2022}

\usepackage{times}
\usepackage{latexsym}

\usepackage[]{todonotes}
\usepackage{booktabs}
\usepackage{color}
\usepackage[T1]{fontenc}
\usepackage{multirow}
\usepackage{multicol}
\usepackage{amsmath}
\usepackage{amssymb}
\usepackage{algorithmic}
\usepackage{times}
\usepackage{latexsym}
\usepackage{graphicx}
\usepackage{url}
\usepackage{algorithm}
\usepackage{amsmath}
\usepackage{amssymb}
\usepackage{multirow}
\usepackage{arydshln}
\usepackage[utf8]{inputenc}

\usepackage{microtype}

\usepackage{inconsolata}

\title{Soft-Labeled Contrastive Pre-training for Function-level 
\mbox{Code Representation}}

\author{
Xiaonan Li\textsuperscript{1}\thanks{\ \ Equal contribution and work is done during the internship at Microsoft Research Asia.}\ , Daya Guo\textsuperscript{2}$^{*}$, Yeyun Gong\textsuperscript{2}, Yun Lin\textsuperscript{3}, Yelong Shen\textsuperscript{2},\\
\textbf{Xipeng Qiu\textsuperscript{1}\thanks{\ \ Corresponding author.}\ , Daxin Jiang\textsuperscript{2}, Weizhu Chen\textsuperscript{2}, Nan Duan\textsuperscript{2}} \\
\textsuperscript{1} Shanghai Key Laboratory of Intelligent Information Processing, Fudan University \\
\textsuperscript{1} School of Computer Science, Fudan University  \textsuperscript{2}Microsoft
\textsuperscript{3}National University of Singapore\\
\textsuperscript{1}\{lixn20, xpqiu\}@fudan.edu.cn,
\textsuperscript{2}\{t-dayaguo, yegong, yeshe, \\ djiang, wzchen, nanduan\}@microsoft.com, 
\textsuperscript{3}dcsliny@nus.edu.sg
}

\begin{document}
\maketitle
\begin{abstract}

Code contrastive pre-training has recently achieved significant progress on code-related tasks. In this paper, we present \textbf{SCodeR}, a \textbf{S}oft-labeled contrastive pre-training framework with two positive sample construction methods to learn functional-level \textbf{Code} \textbf{R}epresentation. Considering the relevance between codes in a large-scale code corpus, the soft-labeled contrastive pre-training can obtain fine-grained soft-labels through an iterative adversarial manner and use them to learn better code representation. 
The positive sample construction is another key for contrastive pre-training. Previous works use transformation-based methods like variable renaming to generate semantically equal positive codes. However, they usually result in the generated code with a highly similar surface form, and thus mislead the model to focus on superficial code structure instead of code semantics. To encourage SCodeR to capture semantic information from the code, we utilize code comments and abstract syntax sub-trees of the code to build positive samples.
We conduct experiments on four code-related tasks over seven datasets. Extensive experimental results show that SCodeR achieves new state-of-the-art performance on all of them, which illustrates the effectiveness of the proposed pre-training method.

\end{abstract}

\section{Introduction}

Function-level code representation learning aims to learn continuous distributed vectors that represent the semantics of code snippets \citep{alon2019code2vec}, which has led to dramatic empirical improvements on a variety of code-related tasks such as code search, clone detection, code summarization, etc. To learn function-level code representation on unlabeled code corpus with self-supervised objectives, recent works \citep{jain2021contrastive,bui2021self,ding2022towards,wang2022code}
propose contrastive pre-training methods for programming language.
In their contrastive pre-training, they usually pull positive code pairs together in representation space and regard different codes as negative pairs via pushing their representation apart. However, they ignore the potential relevance between codes since different programs in a large code corpus may have some similarities. For example, an ascending sort program and a descending sort program are somewhat similar since they both sort their input in a certain order. More seriously, there are a lot of duplications in code corpus~\citep{code_duplication_1,code_duplication_2}, which can cause the “false negative” problem and deteriorate the model~\citep{cv_false_negative_1,cv_false_negative_2}. The other problem of current code contrastive pre-training methods is their positive sample construction.
ContraCode \citep{jain2021contrastive} and Corder \citep{bui2021self} design code transformation algorithms like variable renaming and dead code insertion to generate semantically equivalent programs as positive samples, while Code-MVP \citep{wang2022code} leverages code structures like abstract syntax tree (AST) and control flow graphs (CFG) to transform a program to different variants. 
\begin{figure}[t]
    \centering
    \includegraphics[width=0.95\linewidth]{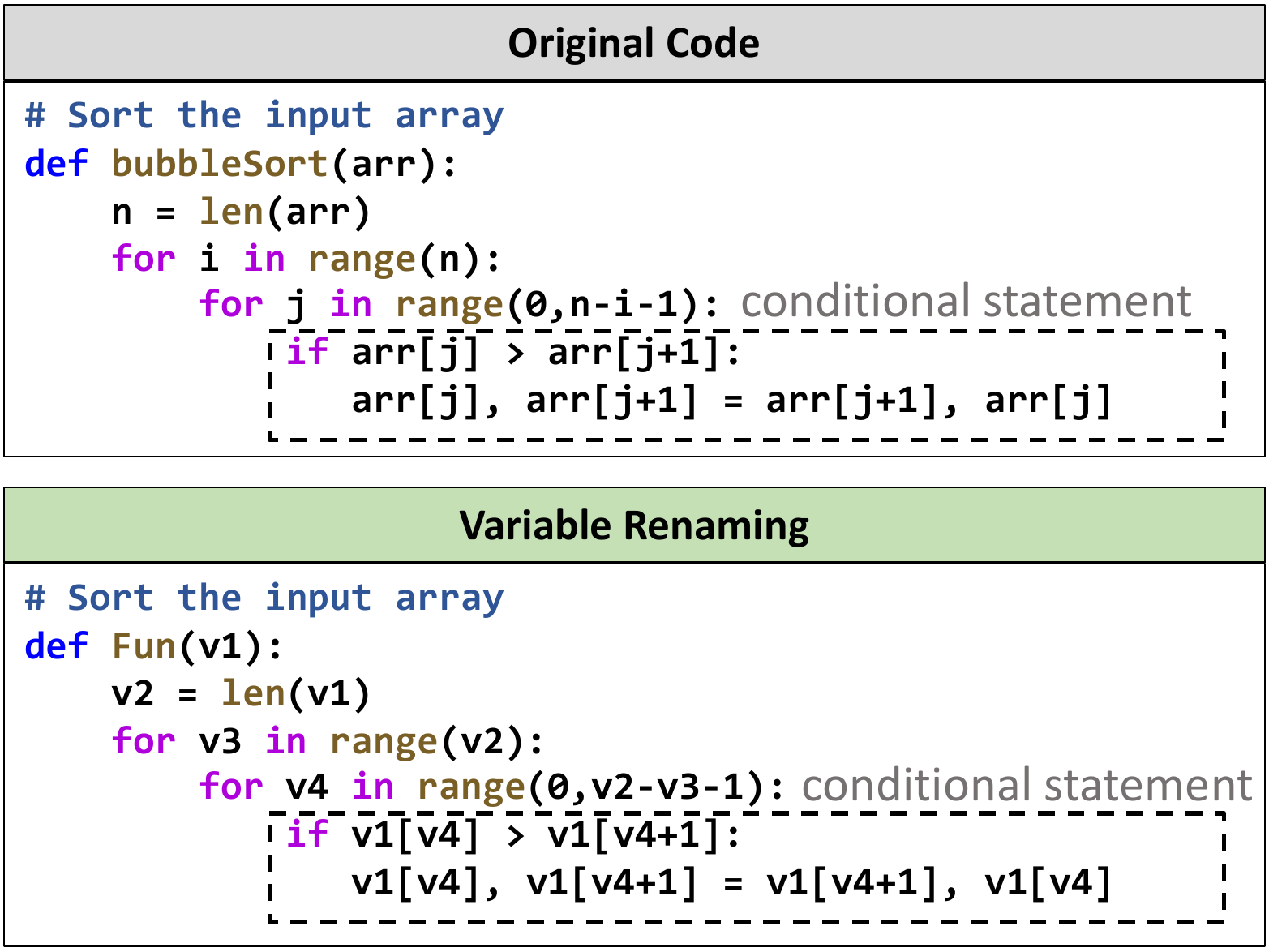}
    \caption{An example of applying variable renaming.}
    \label{fig:code_trans}
    \vskip -0.2in
\end{figure}
However, as shown in Figure \ref{fig:code_trans}, these methods usually result in generated positive samples with highly similar structures (e.g. double loop statements with a conditional statement) to the original program. To pull such positive pairs closer in representation space, the model will tend to learn function-level code representation from superficial code structure rather than substantial code semantics. 
To address these limitations, we present \textbf{SCodeR}, a \textbf{S}oft-labeled contrastive pre-training framework with two positive sample construction methods to learn function-level \textbf{Code} \textbf{R}epresentation.

The soft-labeled contrastive pre-training framework can obtain relevance scores between samples and the original program as soft-labels in an iterative adversarial manner to improve code representation. 
Specifically, we first leverage hard-negative samples from contrastive pre-training to fool discriminators that can explore finer-grained token-level interactions, while discriminators learn to distinguish them and predict relevance scores among samples as soft-labels for contrastive pre-training. 
Through this adversarial iteration, discriminators can provide progressive feedback to improve code contrastive pre-training through soft-labels.

As for positive sample construction, we propose to utilize code comment and abstract syntax sub-tree of the source code to construct positive samples for SCodeR pre-training. Generally, user-written code comments highly describe the function of a source code like “sort the input array” in Figure \ref{fig:code_trans}, which provides crucial semantic information for the model to capture code semantics. 
Besides the comment, the code itself also contains rich information. To further explore the intra-code correlation and contextual knowledge for code contrastive pre-training, we randomly select a piece of code via AST like the conditional statement of Figure \ref{fig:code_trans} and its context as a positive pair. These positive pairs require the model to understand code semantics and learn to infer the selected code based on its context and can help the model learn representation from code semantics.

We evaluate SCodeR on four code-related downstream tasks over seven datasets, including code search, clone detection, zero-shot code-to-code search, and markdown ordering in python notebooks. 
Results show that SCodeR achieves state-of-the-art performance and ablation studies demonstrate the effectiveness of positive sample construction and soft-labeled contrastive pre-training. We release the codes and resources at \url{https://github.com/microsoft/AR2/tree/main/SCodeR}.

\section{Related Works}
\paragraph{Pre-trained Models for Programming Language.}
With the great success of pre-trained models in natural language processing field \citep{devlin2018bert,lewis2019bart,raffel2019exploring,brown2020language}, recent works attempt to apply pre-training techniques on programming languages to facilitate the development of code intelligence. \citet{kanade2019pre} pre-train CuBERT on a large-scale Python corpus using masked language modeling (MLM) and next sentence prediction objectives. \citet{feng2020codebert} pre-train CodeBERT on code-text pairs in six programming languages via MLM and replaced token detection objectives to support text-code related tasks such as code search. GraphCodeBERT \cite{guo2020graphcodebert} leverages data flow as additional semantic information to enhance code representation.
To support code completion, \citet{svyatkovskiy2020intellicode} and \citet{lu2021codexglue} respectively propose GPT-C and CodeGPT. Both of them are decoder-only models and pre-trained by unidirectional language modeling.
Some recent works \citep{ahmad2021unified,wang2021codet5,guo2022unixcoder} explore unified pre-trained models to support both understanding and generation tasks. PLBART \citep{ahmad2021unified} and CodeT5 \citep{wang2021codet5} are based on the encoder-decoder framework. PLBART uses denoising objective to pre-train the model and CodeT5 considers the crucial token type information from identifiers. 
However, these pre-trained models usually result in poor function-level code representation \citep{guo2022unixcoder} due to the anisotropy representation issue \citep{li2020sentence}.
In this work, we mainly investigate how to learn function-level code semantic representations.

\paragraph{Contrastive Pre-training for Code Representation.}
To learn function-level code semantic representation, several attempts have been made to leverage contrastive pre-training on programming languages. ContraCode \citep{jain2021contrastive} and Corder \citep{bui2021self} design transformation algorithms like variable renaming and dead code insertion to generate semantically equivalent programs as positive instances, while \citet{ding2022towards} design structure-guided code transformation algorithms that inject real-world security bugs to build hard negative pairs for contrastive pre-training. Instead of using semantic-preserving program transformations, SynCoBERT \cite{wang2022syncobert} and Code-MVP \citep{wang2022code} construct the positive pairs
through the compilation process of programs like AST and CFG.
However, these works usually generate positive samples with highly similar structures as the original program.
To distinguish these positive samples from candidates,
the model might learn code representation from code surface forms according to hand-written patterns, instead of code semantics.
In this paper, we propose to utilize the comment and abstract syntax sub-tree of the code to construct positive samples and present a method to obtain relevance scores among samples as soft-labels for contrastive pre-training.

\section{Positive Sample Construction}
\label{sec:positive}
In this section, we describe how to construct positive pairs for SCodeR. 
Different from previous works that design transformation algorithms to generate semantically equivalent but highly similar programs, we propose to leverage comment and abstract syntax sub-tree of the code for positive sample construction to encourage the model to capture semantic information.

\subsection{Code Comment}
User-written code comments usually summarize the functionality of the codes and provide crucial semantic information about the source code. Taking the code in Figure~\ref{fig:code_asst} as an example, the comment “sort the input array” highly describes the goal of the code and can help the model to understand code semantics from the natural language. Therefore, we take source code $c$ with the corresponding comment $t$ as positive pair $(t,c)$.
Such positive pairs not only enable the model to understand the code semantics but also align the representation of different programming languages with a unified natural language description as a pivot.

\subsection{Abstract Syntax Sub-Tree}
\begin{figure}
    \centering
    \includegraphics[width=0.99\linewidth]{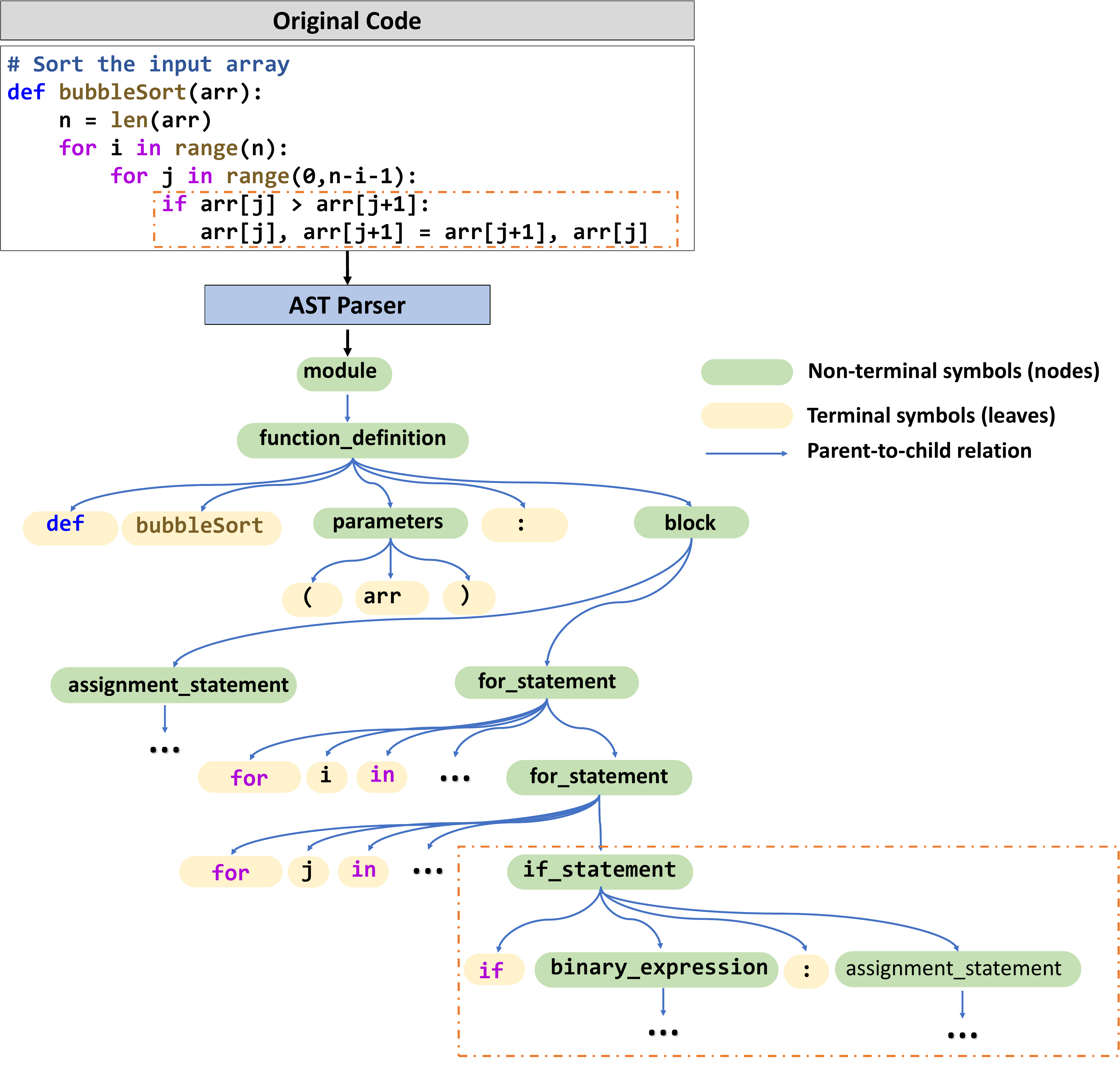}
    \caption{An ASST example of bubble sort.}
    \label{fig:code_asst}

\end{figure}
Besides the comment, the code itself also contains rich information. To further explore the intra-code correlation and contextual knowledge for contrastive pre-training, we propose a method, called \textbf{A}bstract \textbf{S}yntax \textbf{S}ub-\textbf{T}ree Extraction (\textbf{ASST}), that leverages the abstract syntax sub-tree of the source code to construct positive code pairs. 
We give an example of a Python code with its AST in Figure~\ref{fig:code_asst}. We first randomly select the sub-tree of the AST like “if statement”, and then take the corresponding code of the sub-tree and the remaining code as positive code pairs. The procedure of extraction is illustrated in Algorithm~\ref{alg:asst}.
Specifically, we first pre-define a set $N$ of node types whose sub-trees can be used to construct positive pairs. The set mainly consists of statement-level types like “for\_statement” that usually contain a complete and functional code snippet. We then start from a randomly selected leaf node (line 1-2) and find an eligible node in the pre-defined set $N$ along the direction of the root node (line 3-10). Finally, we take the corresponding code $s$ (i.e. leaf children) of the eligible node and the remaining code context $\tilde{s}$ as a positive code pairs $(s,\tilde{s})$ for contrastive pre-training. To avoid extracting those code spans that are too short or meaningless, we set a minimum length $l_{min}$ for the extracted code spans  $s$.

While transformation-based methods generate programs with similar structures, the structures of positive code pairs generated by ASST are different since they belong to different parts of a program. Meanwhile, they are logically relevant because they compose a program of full functionality. To estimate which code is complementary to a given code context in contrastive pre-training, the model needs to understand code semantics and learn to reason based on its context, which encourages the model to understand code semantics.

\begin{algorithm}[t]
\small
	\caption{Abstract Syntax Sub-Tree Extraction}
	\label{alg:asst}
	\begin{algorithmic}[1]
		\REQUIRE The AST $T$ of a code $c$ and pre-defined selectable node types $N$.
		\STATE Collect leaf children $C$ of nodes whose types are in $N$
        \STATE  Randomly sample a node $a$ from $C$
		\WHILE{True} 
		\STATE $s \gets$ the corresponding code of $a$
		\IF{$length(s) \geq l_{min}$ and $a\in N$} 
		\RETURN $s$
		\ELSE
		\STATE $a \gets$ the parent node of $a$
		\ENDIF
		\ENDWHILE 
	\end{algorithmic} 
\end{algorithm}

There are similar mechanisms to learn text representation such as Inversed Cloze Task (ICT) \citep{lee2019latent} that takes a random span of natural language tokens and their context as a positive pair. However, ICT cannot be directly applied to code because code has an explicit structure. If we randomly select code spans on token-level, the selected code spans might be  ungrammatical such as \ \textit{``for i''}, which will mislead the model to focus on structural matching rather than semantic matching.

\section{Soft-Labeled Contrastive Pre-training}
Previous code contrastive pre-training methods usually take different programs in a code corpus as negative pairs and push them apart in the representation space. 
However, different programs in an unlabeled code corpus may have some similarities. 
Taking a program that sorts the input in ascending order as an example, even though another “descendingly sort” program is not semantically equal with it, they both sort their input in a certain order and thus are somewhat similar. Another problem is the “false negative” issue~\citep{cv_false_negative_1,cv_false_negative_2} due to the duplication  in the code corpus~\citep{code_duplication_1,code_duplication_2}. 
To alleviate these problems, we propose soft-labeled contrastive pre-training framework that uses relevance scores between different samples as soft-labels to learn function-level code representation. 

\subsection{Overview}
The soft-labeled contrastive pre-training framework involves three components: (1) A dual-encoder ${G}_{\theta}$ that aims to learn function-level code representation (2) Two discriminators $D_{\phi}$ and $D_{\psi}$ that calculate relevance scores between two inputs for text-code and code-code pairs, respectively.
These components compute the similarity between two samples $(x,y)$ as follows:
\begin{align}
G_{\theta}(x,y) &= E_{\theta}(x)^T {E}_{\theta}(y)
\\
D_{\phi}(x,y) &= \mathbf{w_{\phi}}^T E_{\phi}\left(\left[x;y\right]\right)
\\
D_{\psi}(x,y) &= \mathbf{w_{\psi}}^T E_{\psi}\left(\left[x;y\right]\right)
\end{align}
where $E_{\theta}$, $E_{\phi}$ and $E_{\psi}$ are multi-layer Transformer \citep{vaswani2017attention} encoders with mean-pooling. $w_{\phi}$ ($w_{\psi}$) is a linear layer to obtain similarity score and $[\cdot; \cdot]$ indicates the concatenation operator. If the input $(x,y)$ is a text-code pair, we use $D_{\phi}$ to calculate the similarity, otherwise we use $D_{\psi}$. 

While the dual-encoder encodes samples separately, discriminators take the concatenation of two samples as the input and fully explore finer-grained token-level interactions through the self-attention mechanism, which can predict more accurate relevance scores between two samples. Therefore, we propose to utilize relevance scores from discriminators as soft-labels to help the encoder $E_{\theta}$ learn better code representation. 

\begin{algorithm}[t]
\small
	\caption{Soft-Labeled contrastive pre-training}
	\label{alg:scl}
	\begin{algorithmic}[1]
		\REQUIRE A dual-encoder $G_\theta$, two discriminators $D_{\phi(\psi)}$, and a set $X$ of positive pairs with a unlabeled code corpus $C$.
		\STATE Initialize the dual-encoder and discriminators.
        \STATE Train the warm-up dual-encoder.
        \STATE Get top-$K$ negative codes $C^{x_k}_{hard}$ from  $C$ for each positive pair $(x_k,x_{k}^+) \in X$ using the dual-encoder.
		\FOR{$i$ in $1\cdots I$} 
		
		\FOR{Discriminators training step}
		\STATE Sample hard negative codes from $C^{x_k}_{hard}$.
		\STATE Update parameters of discriminators $D_\phi$ and $D_\psi$. 
		\ENDFOR
		
		\FOR{Dual-encoder training step}
        \STATE Sample hard negative codes from $C^{x_k}_{hard}$ and obtain relevance scores from discriminators.
        \STATE Update parameters of the dual-encoder $G_\theta$.
		\ENDFOR
		\STATE Refresh Top-$K$ negative codes $C^{x_k}_{hard}$ using new $G_\theta$.

		\ENDFOR
	\end{algorithmic} 
\end{algorithm}
We show the detailed illustration of our proposed soft-labeled contrastive pre-training in Algorithm~\ref{alg:scl}. Specifically,  we first initialize all encoders with a pre-trained model like UniXcoder \citep{guo2022unixcoder} and follow \citet{code_retriever} to train a warm-up dual-encoder using a simple strategy where negative samples come from other positive pairs in the same batch $\mathbb{X}_b$ (line 1-2 of Algorithm \ref{alg:scl}). The loss is calculated as follows, 
\begin{equation}
p_\theta(x^+|x,\mathbb{X}_b)=\frac{e^{G_{\theta}(x,x^+)}}{\sum_{x' \in  \mathbb{X}_b}e^{G_{\theta}(x,x')}}
\end{equation}
\begin{equation}
L_{warm}^{\theta}=-\log p_\theta(x^+|x,\mathbb{X}_b),
\end{equation}
where $(x,x^+) \in X$ is a positive pair as described by Section \ref{sec:positive}.

We then iteratively alternate two training procedures: (1) The dual-encoder is used to obtain hard-negative codes to train the discriminators (line 5-8). (2) The optimized discriminators predict relevance scores among samples as soft-labels to improve the dual-encoder (line 9-12). Through this iterative training, the dual-encoder gradually produces harder negative samples to train better discriminators, whereas the discriminators provide better progressive feedback to improve the dual-encoder. The details about training procedures for the discriminators and dual-encoder will be described next.

\subsection{Discriminators Training}
\label{sec:train_cross_encoder}
Given a text $x$ from positive text-code pairs $(x,x^+)$, the discriminator $D_{\phi}$ is optimized by maximizing the log likelihood of selecting  positive code $x^+$ from candidates $\mathbb{X}$ as follows,
\begin{equation}
p_\phi(x^+|x,\mathbb{X})=\frac{e^{D_{\phi}(x,x^+)}}{\sum_{x' \in  \mathbb{X}}e^{D_{\phi}(x,x')}}
\end{equation}
\begin{equation}
L^\phi=-log p_\phi(x^+|x,\mathbb{X}),
\end{equation}
where $\mathbb{X}$ is the set of negative codes $\mathbb{X}^-$ with a positive code $x^+$. If $x$ is a code from positive code-code pairs, the calculation of $p_\psi$ and $L^\psi$ are analogous to $p_\phi$ and $L^\phi$, respectively.

To better train discriminators, we take those hard-negative examples that are not positive samples but closed to the original example $x$ in the vector space as the negative candidates $\mathbb{X}^-$. In practice, we first get the top-K code samples that are closest to $x$  using $G_\theta$ as the distance function and randomly sample examples from them to obtain a subset $\mathbb{X}^-$.

\subsection{Dual-Encoder Training}

After training discriminators, we utilize relevance scores predicted by discriminators as soft-labels and follow \citet{ar2} to use adversarial and distillation losses to optimize the dual-encoder.

\begin{table*}[t]
\small
\centering
\begin{tabular}{@{}l|ccccccc|c|c@{}}
\toprule
\textbf{Dataset}       & \multicolumn{7}{c|}{\textbf{CSN}}                               & \textbf{AdvTest} & \textbf{CosQA} \\ \midrule
\textbf{Lang}          & \textbf{Ruby} & \textbf{Javascript} & \textbf{Go} & \textbf{Python} & \textbf{Java} & \textbf{PHP} & \textbf{Average} & \textbf{Python}  & \textbf{Python} \\ \midrule
CodeBERT
& 67.9                & 62.0                & 88.2            & 67.2            & 67.6            & 62.8           &69.3 & 27.2&64.7 \\
GraphCodeBERT
& 70.3                & 64.4                & 89.7            & 69.2            & 69.1            & 64.9           &71.3 & 35.2 & 67.5\\
SyncoBERT
& 72.2                & 67.7                & 91.3            & 72.4            & 72.3            & 67.8           &74.0 &38.1 & - \\
CodeRetriever
& 75.3               & 69.5               & 91.6            & 73.3           & 74.0           & 68.2         & 75.3 & 43.0 & 69.6\\
Code-MVP
&- &- &- &- &- &- &- &40.4 &72.1 \\
UniXcoder
&74.0 & 68.4 &91.5 &72.0 & 72.6 & 67.6 & 74.4 & 41.3 & 70.1 \\
SCodeR & \textbf{77.5} & \textbf{72.0} & \textbf{92.7} & \textbf{74.2} & \textbf{74.8} & \textbf{69.2} & \textbf{76.7} & \textbf{45.5} & \textbf{74.5}\\
\bottomrule

\end{tabular}
\vspace{-5pt}
\caption{The comparison on code search task. The results of compared models are from their original papers.}
\label{result_code_search}
\vspace{-10pt}
\end{table*}

\paragraph{Adversarial loss:} 
\begin{equation}
\label{eq_adv}
{L}^{\theta}_{adv}=-\sum_{x^-\in \mathbb{X}^-}w(x^-)* \log p_{\theta}(x^- |  x, \mathbb{X}^- )
\end{equation}
where $w(x^-)$ is 
$-\log p_\phi( x^+|x,\{x^+,x^-\})$ if $x$ is a text otherwise $w$ is $-\log p_\psi( x^+|x,\{x^+,x^-\})$. We apply the same approach to obtain hard-negative candidates $\mathbb{X}^-$ as described in Section~\ref{sec:train_cross_encoder}.

When optimizing $G_\theta$, $w$ in Equation~\ref{eq_adv} is a constant and adjusts weight for each negative example. 
When $-\log p_{\phi(\psi)}( x^+|x,\{x^+,x^-\})$ is small, i.e. discriminators predict that $x$ and $x^-$ are semantically relevant, $w$ will be a high weight and force $G_\theta$ to draw the representation of $x$ and $x^-$ closer among $\mathbb{X}^-$.
Since we optimize the dual-encoder on negative codes under different weight $w$, the representation of negative codes with high relevance score will be closer to $x$, and those with low relevance score will be pushed away.

\paragraph{Distillation loss:} 
\begin{equation}
\begin{aligned}
    L^{\theta}_{distill} & = H(p_{\phi(\psi)}(\cdot |x,\mathbb{X}), p_{\theta}(\cdot | x,\mathbb{X}) )
\end{aligned}
    \label{eq_distill}
\end{equation}

We also use a distillation loss function~\citep{knowledge_distillation_hinton} to encourage the dual-encoder to fit the probability distribution of discriminators over $\{x^+\}\cup\mathbb{X}^-$ using KL divergence loss $H$.
Through $L^{\theta}_{distill}$, we can inject discriminators' knowledge into the dual-encoder by soft-labels $p_{\phi(\psi)}$. 

\paragraph{Training Objective of Dual-Encoder}
The overall loss function of the dual-encoder is the integration of adversarial loss and distillation loss as follows, where $\lambda$ is a pre-defined hyper-parameter.
\begin{equation}
    L^\theta = \lambda * L^{\theta}_{adv} + (1-\lambda) * L^{\theta}_{distill}
\end{equation}
Through $L_\theta$, we can provide discriminators' progressive feedback to the dual-encoder through soft-labels.
After this adversarial iteration, we will use $E_\theta$ to serve for downstream tasks.

\section{Experiment}

\subsection{Model Comparison}
We compare SCodeR with various state-of-the-art pre-trained models. \textbf{RoBERTa}~\citep{liu2019roberta} is pre-trained on text corpus by masked language model (MLM). \textbf{CodeBERT}~\citep{feng2020codebert} is pre-trained on large scale code corpus with MLM and replaced token detection. \textbf{GraphCodeBERT}~\citep{guo2020graphcodebert} is based on CodeBERT and integrates the data flow information to enhance code representation. \textbf{PLBART}~\citep{ahmad2021unified} is adapted from the BART~\citep{lewis2019bart} architecture and pre-trained using denoising objective on Java, Python and stackoverflow corpus. \textbf{CodeT5}~\citep{wang2021codet5} is based on the T5~\citep{nl_t5} architecture, considering the identifier token information and applying multi-task learning. \textbf{UniXcoder}~\citep{guo2022unixcoder} is adapted from the UniLM~\citep{unilm} architecture, pretrained by different tasks (understanding and generation) on unified cross-modal data (code, AST and text). We also compare SCodeR with those code pre-trained models that utilize contrastive pre-training. \textbf{SynCoBERT}~\citep{wang2022syncobert} and \textbf{Code-MVP}~\citep{wang2022code} construct positive pairs through multiple views of code like AST and CFG. \textbf{Corder}~\citep{bui2021self} and \textbf{DISCO}~\citep{ding2022towards} construct positive code pairs from semantic-preserving transformations, and the latter additionally uses bug-injected codes as hard negatives. \textbf{CodeRetriever}~\citep{code_retriever} builds code-code pairs by corresponding documents and function name automatically. 
For fair comparison, we use the same model architecture, pre-training corpus, and downstream hyper-parameters as previous works~\citep{code_retriever,guo2022unixcoder}. To accelerate the training process, we initialize dual-encoder and discriminators with the released parameters of UniXcoder~\citep{guo2022unixcoder}.
More details about pre-training and fine-tuning can be found in the Appendix~\ref{sec:appendix_pretrain_hyper_parameters} and \ref{sec:appendix_finetuning_hyper_parameters}.

\subsection{Natural Language Code Search}

Given a natural language query as the input, code search aims to retrieve the most semantically relevant code from a collection of code candidates. We conduct experiments on CSN~\citep{guo2020graphcodebert}, AdvTest~\citep{lu2021codexglue} and CosQA~\citep{huang2021cosqa} to evaluate SCodeR. CSN contains six programming languages, including Ruby, Javascript, Python, Java, PHP and Go. The dataset is constructed from CodeSearchNet Dataset~\citep{husain2019codesearchnet} and noisy queries with low quality are filtered. AdvTest normalizes the function name and variable name of python code and thus is more challenging. The queries of CosQA are from Microsoft Bing search engine, which makes it closer to real-world code search scenario. Following previous works~\citep{feng2020codebert,guo2020graphcodebert,guo2022unixcoder}, we adopt Mean Reciprocal Rank (MRR)~\citep{mrr1} as the evaluation metric. 

The results are shown in Table~\ref{result_code_search}. We can see that SCodeR outperforms previous code pre-trained models and achieves the new state-of-the-art performance on all datasets. 
Specifically, SCodeR outperforms UniXcoder by 2.3 points on the CSN dataset, and improves over state-of-the-art models about 2.5 points on AdvTest and CosQA datasets, which demonstrates the effectiveness of SCodeR.

\subsection{Code Clone Detection}
\begin{table}[h]
\centering
\small
\setlength{\tabcolsep}{0.5mm}
\begin{tabular}{@{}lcccc@{}}
\toprule
              & \multicolumn{1}{l}{\textbf{POJ-104}} & \multicolumn{3}{c}{\textbf{BigCloneBench}}                                                         \\ \midrule
                          & \multicolumn{1}{l}{MAP@R}   & \multicolumn{1}{l}{Recall} & \multicolumn{1}{l}{Precision} & \multicolumn{1}{l}{F1-score} \\
RoBERTa       & 76.67                       & 95.1                       & 87.8                          & 91.3                         \\
CodeBERT      & 82.67                       & 94.7                       & 93.4                          & 94.1                         \\
GraphCodeBERT & 85.16                       & 94.8                       & 95.2                          & 95.0                         \\
SyncoBERT     & 88.24                       & -                            &-                               &-                              \\
CodeRetriever & 88.85 & -&- &- \\
Corder &84.10 &- &- &- \\
DISCO &82.77 &94.6 &94.2 &94.4 \\
PLBART        & 86.27                       & 94.8                       & 92.5                          & 93.6                         \\
CodeT5-base   & 88.65                       & 94.8                       & 94.7                          & 95.0                         \\
UniXcoder     & 90.52                       & 92.9                       & \textbf{97.6}                          & 95.2                         \\
SCodeR    & \textbf{92.45}                        & \textbf{96.2}                      & 94.5                         & \textbf{95.3}                     \\ \bottomrule
\end{tabular}
\caption{Performance on code clone detection. The results of compared models are from their original papers.}
\label{result_code_clone}
\end{table}
Code clone detection aims to identify the semantic similarity between two codes. We consider POJ-104~\citep{poj_104} and BigCloneBench~\citep{bcb} to evaluate SCodeR. POJ-104 dataset (C/C++) consists of codes from online judge (OJ) system. It aims to find the semantically similar codes given a code as query and evaluates by Mean Average Precision (MAP). BigCloneBench dataset (Java) is to judge whether two codes are similar and evaluates by Precision, Recall, and F1-score.
We show the results in Table~\ref{result_code_clone}. 

Compared with  previous pre-trained models, SCodeR achieves the overall best performance on both datasets. On POJ-104 dataset, SCodeR surpasses all other methods. Specifically, SCodeR outperforms UniXcoder by 1.93 points. Although the pre-training corpus does not cover C/C++ programming languages, the superior performance reflects that SCodeR learns better general code knowledge.
On the BigCloneBench dataset, SCodeR also achieves comparable performance.
These results show that SCodeR learns better function-level code representation for code clone detection.

\subsection{Zero-Shot Code-to-Code Search}
\begin{table*}[t]
\small
\centering
\begin{tabular}{@{}l|ccc|ccc|ccc|c@{}}
\toprule
\textbf{Query PL}      & \multicolumn{1}{l}{}     & \multicolumn{1}{l}{\textbf{Ruby}}   & \multicolumn{1}{l|}{}     & \multicolumn{1}{l}{}     & \multicolumn{1}{l}{\textbf{Python}} & \multicolumn{1}{l|}{}     & \multicolumn{1}{l}{}     & \multicolumn{1}{l}{\textbf{Java}}   & \multicolumn{1}{l|}{}     & \multirow{2}{*}{\textbf{Overall}} \\ \cmidrule(r){1-10}
\textbf{Target PL}     & \multicolumn{1}{l}{\textbf{Ruby}} & \multicolumn{1}{l}{\textbf{Python}} & \multicolumn{1}{l|}{\textbf{Java}} & \multicolumn{1}{l}{\textbf{Ruby}} & \multicolumn{1}{l}{\textbf{Python}} & \multicolumn{1}{l|}{\textbf{Java}} & \multicolumn{1}{l}{\textbf{Ruby}} & \multicolumn{1}{l}{\textbf{Python}} & \multicolumn{1}{l|}{\textbf{Java}} &                          \\ \midrule
CodeBERT & 13.55 & 3.18 & 0.71 & 3.12 & 14.39 & 0.96 & 0.55& 0.42& 7.62& 4.94 \\
GraphCodeBERT & 17.01 & 9.29 & 6.38 & 5.01 & 19.34 & 6.92 & 1.77& 3.50& 13.31& 9.17\\
PLBART & 18.60 & 10.76 & 1.90 & 8.27 & 19.55 & 1.98 & 1.47& 1.27& 10.41& 8.25 \\
CodeT5-base & 18.22 & 10.02 & 1.81 & 8.74 & 17.83 & 1.58 & 1.13& 0.81& 10.18& 7.81 \\
UniXcoder& 29.05 & 26.36 & 15.16 & 23.96 & 30.15 & 15.07 & 13.61& 14.53& 16.12& 20.45 \\
SCodeR& \bf{33.87} & \bf{30.25} & \bf{17.10} & \bf{26.48} & \bf{33.02} & \bf{16.95} & \bf{16.5}& \bf{19.06}& \bf{18.87}& \bf{23.57} \\\bottomrule
\end{tabular}
\caption{The comparison on zero-shot code-to-code search. Baselines' results are reported by \citet{guo2022unixcoder}.}
	\label{result_zero_shot_code_to_code_search}
	\vspace{-10pt}
\end{table*}

We also evaluate SCodeR in zero-shot code-to-code search. Given a code snippet as query, the task aims to find semantically similar codes from a collection of code candidates in zero-shot setting. Since the annotation of code-to-code search is labor-intensive and costly~\citep{bigclonebench,code_retriever}, the zero-shot performance can indicate the model's utility in real-world scenario, where a lot of programming languages do not have an annotated dataset for code-to-code search.
We follow \citet{guo2022unixcoder} to conduct the experiment on CodeNet~\citep{codenet} and evaluate models using  MAP score. 
The results are listed in Table~\ref{result_zero_shot_code_to_code_search}. The first and the second row correspond to query and target programming languages.

We can see that SCodeR outperforms all other compared models and improves over the state-of-the-art model, i.e. UniXcoder, by 3.12 average absolute points. Meanwhile, SCodeR has a consistent improvement on the cross-PL setting, which can help users to translate programs from one PL to another via retrieving semantically relevant codes.

\begin{table}[t]
\centering
\small
\begin{tabular}{@{}lc@{}}
\toprule
Model         & Kendall's Tau \\ \midrule
CodeBERT~\citep{feng2020codebert}     & 81.9       \\
GraphCodeBERT~\citep{guo2020graphcodebert}\hspace{-15pt} & 84.7       \\
PLBART~\citep{ahmad2021unified}        & 84.7       \\
CodeT5~\citep{wang2021codet5}        & 84.7       \\
UniXcoder~\citep{guo2022unixcoder}     & 85.9       \\
SCodeR    & \textbf{86.6}       \\ \bottomrule
\end{tabular}

\caption{Experiment results on markdown ordering in python notebooks.}
\label{table:result_notebook_reorder}
\end{table}

\begin{table*}[h]
	\begin{center}
		\begin{small}
			\begin{tabular}{lccccccc}
				\toprule
				 Methods & Ruby & Javascript & Go & Python & Java & Php & Overall\\
				\midrule
				Baseline    &74.0 & 68.4 &91.5 &72.0 & 72.6 & 67.6 & 74.4 \\	
				\hdashline
				Baseline + Code Transformation & 74.5 & 68.7 & 91.9 & 72.2 & 72.6 & 67.7 & 74.6 \\
				Baseline + ASST & 76.1 & 70.1 & 92.1 & 73.0 &73.3 &68.1&75.4 \\
				\hdashline
				Baseline + ASST + Code Comment & 76.2 &71.2 &92.2&73.4&73.7&68.5&75.9\\
				Baseline + ASST + Code Comment + Soft-Labled & \textbf{77.5}&\textbf{72.0}&\textbf{92.7}&\textbf{74.2}&\textbf{74.9}&\textbf{69.2}&\textbf{76.8}\\

				\bottomrule
			\end{tabular}
        	\caption{Ablation study on natural language code search.}
	\label{ablation_stdudy_code_search}
	\vspace{-10pt}
		\end{small}
	\end{center}

\end{table*}

\subsection{Markdown Ordering in Python Notebooks}
This task is to reconstruct the order of markdown cells in a given notebook according to the ordered code cells. We conduct experiments on the dataset provided by Kaggle\footnote{https://www.kaggle.com/competitions/AI4Code/overview} and use the official evaluation metric, \textit{Kendall's tau ($\tau$)}.  It is computed as $1-2*N/\tbinom{n}{r}$ where $N$ is the number of pairs
in the predicted sequence with incorrect relative order and $n$ is the sequence length.

We take the normalized markdown cell's position in a given notebook as labels for each markdown cell (0$\sim$1), and solve this task as a regression task. To test performance of function-level code representation, we  use pre-trained models to encode each cell to function-level representation as features.  We use a randomly initialized Transformer that takes extracted features of cells in the python notebook to predict position of each cell. Note that parameters of pre-trained models are fixed in the fine-tuning procedure, and thus the performance of this task depends on function-level feature extracted from pre-trained models.

We show the results in Table~\ref{table:result_notebook_reorder}. SCodeR outperforms other pre-trained models and achieves 0.5 points higher than UniXcoder. This indicates that SCodeR learns better representation for both code and natural language comments, and can help better understand the fine-grained relationship of codes and comments in the python notebook.

\subsection{Analysis}

\paragraph{Ablation Study}

To evaluate the effect of our positive sample construction methods and soft-labeled contrastive pre-training framework, we conduct ablation study on the CSN dataset and take the pre-trained model with no enhancement as the baseline (i.e. UniXcoder). At first, we individually compare the proposed ASST with the transformation-based positive sample construction method~\citep{contracode,bui2021self}.
Notice that previous works do not apply their transformation-based methods on all six programming languages covered by our pre-training corpus. For fair comparison and keeping the pre-training corpus consistent, we follow~\citet{reacc} to implement the widely used transformations including variable renaming and dead code insertion on six programming languages by ourselves. 
Then, we add the remaining modules of SCodeR to evaluate their performance. The results are shown in Table~\ref{ablation_stdudy_code_search}. 

Compared with transformation-based methods, we can see that our positive sample construction (ASST) achieves better performance.
Meanwhile, positive pairs from ASST can bring significant improvement over the baseline, which reflects its effectiveness. After using the text-code pairs, the performance improves over 0.5 points, which shows that code comments provide rich semantic information to  help model learn better code representation. When adding soft-labeled contrastive pre-training, the model performance increases by 0.9 points, which demonstrates that applying relevance among samples as soft-labels for contrastive learning can further improve code representation.

\paragraph{Effect of AST-based Splitting}
\begin{table}[h]
\centering
\small
\begin{tabular}{@{}lcccc@{}}
\toprule
              & \textbf{Ruby}  & \textbf{Python} & \textbf{Java}  & \textbf{Overall} \\ \midrule
SCodeR & \textbf{33.87} & \textbf{33.02} &\textbf{18.87} &\textbf{28.6}\\
SCodeR$_T$ & 32.97 & 30.78 & 18.01 & 27.2 \\
SCodeR$_L$ & 33.22 & 31.31  & 18.23 & 27.6    \\ \bottomrule
\end{tabular}
\caption{Experiment results of different strategies of code splitting on zero-shot code-to-code search. SCodeR$_T$ and SCodeR$_L$ use token-level and line-level ICT to replace ASST.}
\label{table:different_code_splitting}
\end{table}
We conduct experiments on zero-shot code-to-code search to analyze the effect of AST-based splitting strategy of ASST by comparing ASST with two variants of splitting strategy. The first strategy is token-level ICT that takes a random span of code tokens and their context as positive pairs. The second strategy is line-level ICT that considers random consecutive code lines and the remaining lines as positive pairs.  Compared with our AST-based splitting method, these two splitting strategies will cause ungrammatical codes and mislead the model to focus on structural matching rather than semantic matching. The results are shown in Table~\ref{table:different_code_splitting} and we can see that the two variants of splitting strategy lead to worse performance, which shows the effectiveness of our AST-based splitting method.

\paragraph{Case Study}
\begin{figure}
    \centering
    \includegraphics[width=0.9\linewidth]{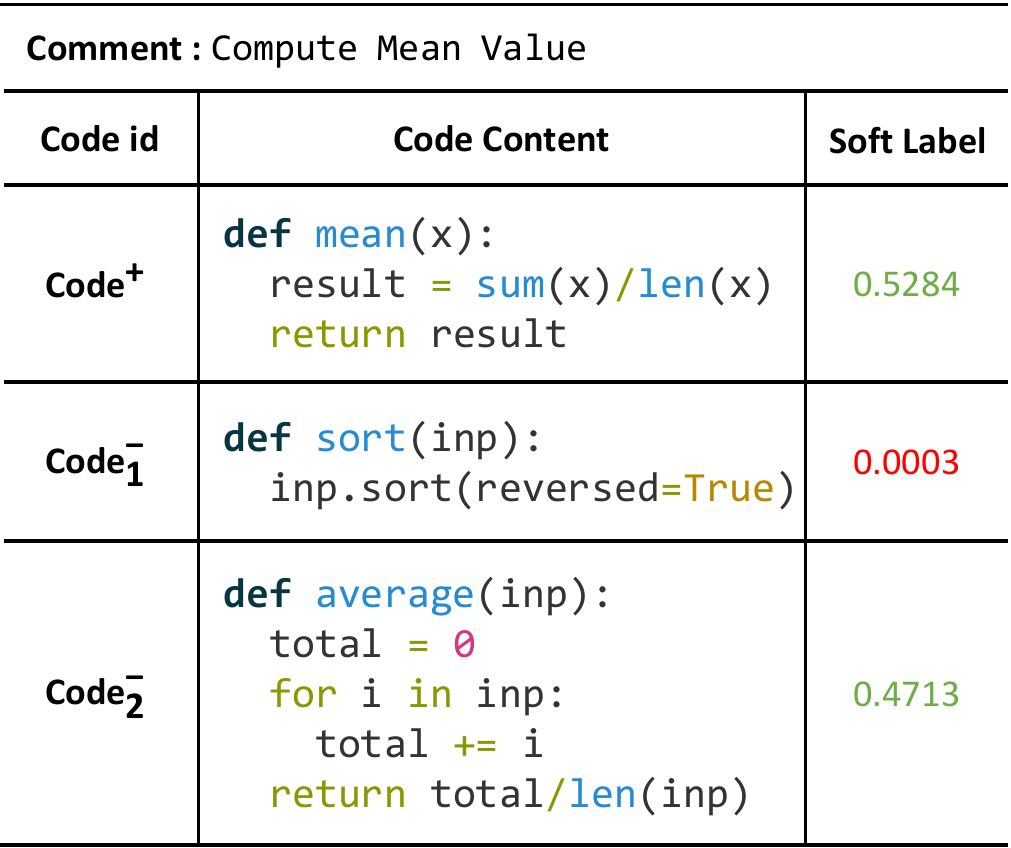}
    \caption{Case study on the discriminator. The soft-label is the relevance scores between the comment and codes from the discriminator, $p_{\phi}(\cdot |x,\mathbb{X})$.}
    \label{fig:case_study}
    \vskip -0.2in
\end{figure}
We give a case study in Figure~\ref{fig:case_study} to show the importance of soft-labels for contrastive pre-training. The figure includes one paired code and two other codes with soft-labels provided by the discriminators. We can see that the soft-label of negative $\text{Code}^-_1$ is close to 0 since the code is unrelated with the comment and the discriminators can predict correct relevance score between them for contrastive pre-training. 
$\text{Code}^-_2$ is a false negative that has the same functionality as $\text{Code}^+$ and should be assigned similar weights when we apply contrastive pre-training. As we can see in the figure, the discriminator can understand code semantics and provide similar soft-labels (i.e 0.5284 vs 0.4713) about $\text{Code}^+$ and $\text{Code}^-_2$ for contrastive pre-training, which can alleviate the influence of false negative issue and learn better code representation through soft-labels.

\section{Conclusion}
In this paper, we present SCodeR to learn function-level code representation with soft-labeled contrastive pre-training. To alleviate the “false negative” issue in code corpus, we propose a soft-labeled contrastive pre-training framework that takes relevance scores among samples as soft-labels for contrastive pre-training in an iterative adversarial manner. Besides, we propose to utilize code comment and abstract syntax sub-tree of the source code to build positive samples that can facilitate the model to capture semantic information from the source code. Experimental results show that SCodeR achieves state-of-the-art performance on four code-related tasks over seven datasets. Further ablation studies show the effectiveness of our soft-labeled contrastive pre-training framework and positive sample construction methods.
\section*{Limitations}
There are two limitations of this work:
1) In the adversarial iteration, we introduce discriminators to provide soft-labels for the training of dual-encoder, which increases GPU memory occupation. To solve it, we can obtain these soft-labels offline, which may complicate the pipeline of data processing.
2) We only use UniXcoder as the backbone model in the experiments due to the computation resources limitation.
We leave pre-training based on other code pre-trained models like CodeBERT~\citep{feng2020codebert}, GraphCodeBERT~\citep{guo2020graphcodebert} and Codex~\citep{codex} as future work.

\bibliography{anthology,custom}
\bibliographystyle{acl_natbib}
\clearpage
\appendix

\section{Pre-training Settings}
\label{sec:appendix_pretrain_hyper_parameters}
For fair comparison, we adopt the same model architecture and the same pre-training corpus as previous works \citep{feng2020codebert,guo2020graphcodebert}. The used corpus is CodeSerachNet~\citep{husain2019codesearchnet}, which includes 2.3M functions paired with documents in six programming languages. We leverage tree-sitter\footnote{https://github.com/tree-sitter/tree-sitter} to get the AST information for ASST. The node set $N$ for ASST includes: “for\_statement”, “while\_statement”, “if\_statement”, “with\_statement”, “try\_statement”, “assignment\_statement”, etc. We also consider the “function\_call” node if it is not under an indivisible node like “assignment\_statement”.
The dual-encoder consists of 12 layers transformer with 768 hidden dimensional hidden states and 12 attention heads. The architecture of discriminators is the same as dual-encoder. To accelerate the training process, we adopt the released parameters of UniXcoder~\citep{guo2022unixcoder} to initialize the dual-encoder and discriminators. 
ScodeR is trained on 8 Nvidia Tesla A100 with 40GB memory and costs about 37 hours. We show the pre-training hyper-parameters in Table~\ref{table:pretrain_hyper}.

\begin{table}[h]
\small
\centering
\begin{tabular}{@{}lcc@{}}
\toprule
Hyper-Parameters            & Dual-encoder             & Discriminators             \\ \midrule
Initialization              & UniXcoder & UniXcoder \\
Optimizer                   & AdamW         & AdamW         \\
Scheduler                   & Linear        & Linear        \\
Warmup proportion           & 0.1           & 0.1           \\
Negative size               & 7             & 7             \\
Batch size                  & 64           & 64           \\
Learning rate               & 5e-6          & 1e-5          \\
Max step                    & 24000         & 16000          \\ 
Iterations & 4 & 4 \\
Loss Weight $\lambda$ &0.2 &-\\
\bottomrule

\end{tabular}
\caption{The hyper-parameters of pre-training.}
\label{table:pretrain_hyper}
\end{table}

\section{Fine-tuning Settings}
\label{sec:appendix_finetuning_hyper_parameters}
\subsection{Natural Language Code Search}
Given a natural language query as the input, code search aims to retrieve the most semantically relevant code from a collection of code candidates. We conduct experiments on CSN~\citep{guo2020graphcodebert}, AdvTest~\citep{lu2021codexglue} and CosQA~\citep{huang2021cosqa} to evaluate SCodeR. 

On CSN, we follow \citet{code_retriever} to set the batch size as 128, learning rate as 2e-5, and max sequence length of PL and NL as 256 and 128. We finetune the model for 10 epochs using AdamW optimzier and select the best checkpoint based on the development set.

On AdvTest dataset, we finetune the model for 2 epochs and keep other hyper-parameters same as CSN dataset.

On CosQA dataset, we use the same hyper-parameters as CSN dataset.
\subsection{Code Clone Detection}
Code clone detection aims to identify the semantic similarity between two codes. We consider POJ-104~\citep{poj_104} and BigCloneBench~\citep{bcb} to evaluate SCodeR. 

On POJ-104 dataset, we follow \citet{guo2022unixcoder} to set the batch size as 8, the learning rate as 2e-5, and the max sequence length as 400. We finetune the model using AdamW optimzier for 2 epochs.

On BigCloneBench dataset, we follow \citet{guo2022unixcoder} to set the batch size as 16, learning rate as 5er-5 and the max sequence length as 512. We use AdamW optimizer to fine-tune the model and select the best checkpoint based on the development set.
\subsection{Markdown Ordering in Python Notebooks}
This task is to reconstruct the order of markdown cells in a given notebook according to the ordered code cells. We conduct experiments on the dataset provided by Kaggle. we use pre-trained models to encode each cell to function-level representation as features and set the max sequence length of each cell as 128. We use a randomly initialized Transformer that takes extracted features of cells in the python notebook to predict position of each cell. We set this Transformer's layers, hidden size, and attention heads as 6, 768, and 12, respectively. For training it, we set the batch size as 128, learning rate as 2e-5, the max number of cells as 256, and the optimizer as AdamW.

\end{document}